\pdfoutput=1

\documentclass[11pt]{article}

\usepackage{ACL2023}
\usepackage{enumerate}
\usepackage{times}
\usepackage{latexsym}
\usepackage{graphicx}
\usepackage{float}
\usepackage{xcolor}
\usepackage{colortbl}
\usepackage{amsmath}
\usepackage{subfigure}
\usepackage{soul}
\usepackage{color}
\usepackage[T1]{fontenc}

\usepackage[utf8]{inputenc}
\usepackage{multirow}
\usepackage{microtype}
\usepackage{inconsolata}
\usepackage{threeparttable}

\title{ Humanity in AI: Detecting the Personality of Large Language Models
}

\author{Baohua Zhang, Yongyi Huang, Wenyao Cui, Huaping Zhang\thanks{*Corresponding author} \and Jianyun Shang \\
       Beijing Institute of Technology, China \\
       kevinzhang@bit.edu.cn
       }

\begin{document}
\maketitle

\begin{abstract}

Questionnaires are a common method for detecting the personality of Large Language Models (LLMs). However, their reliability is often compromised by two main issues: hallucinations (where LLMs produce inaccurate or irrelevant responses) and the sensitivity of responses to the order of the presented options.
To address these issues, we propose combining text mining with questionnaires method.
Text mining can extract psychological features from the LLMs' responses without being affected by the order of options. Furthermore, because this method does not rely on specific answers, it reduces the influence of hallucinations. By normalizing the scores from both methods and calculating the root mean square error, our experiment results confirm the effectiveness of this approach. To further investigate the origins of personality traits in LLMs, we conduct experiments on both pre-trained language models (PLMs), such as BERT and GPT, as well as conversational models (ChatLLMs), such as ChatGPT.
The results show that LLMs do contain certain personalities, for example, ChatGPT and ChatGLM exhibit the personality traits of 'Conscientiousness'. Additionally, we find that the personalities of LLMs are derived from their pre-trained data. The instruction data used to train ChatLLMs can enhance the generation of data containing personalities and expose their hidden personality.
We compare the results with the human average personality score, and we find that the personality of FLAN-T5 in PLMs and ChatGPT in ChatLLMs is more similar to that of a human, with score differences of 0.34 and 0.22, respectively.


\end{abstract}

\section{Introduction}

Large Language Models (LLMs) serve as human assistants that can understand and respond to human language more naturally, help customer service agents respond to client queries promptly and accurately, and offer more personalized experiences~\citep{jeon2023large,liu2023summary,dillion2023can}. Unlike traditional deep learning models, LLMs achieve remarkable performance in semantic understanding and instructions following ~\citep{lund2023ChatGPT,liu2023summary}, which makes LLMs behave more like humans.  

Some research suggests that LLMs are similar to humans in terms of their thinking. For example, \citet{kosinski2023theory} shows that ChatGPT has reached the level of a human 9-year-old child. Additionally, \citet{bubeck2023sparks} demonstrates that  GPT-4 possesses fundamental human-like capabilities. These capabilities include reasoning, planning, problem-solving, abstract thinking, understanding complex ideas, rapid learning, and experiential learning. Experts from Johns Hopkins University have found that the theory of mind of GPT-4 has surpassed human abilities. It achieves 100\% accuracy in some tests through a process of mental chain reasoning and step-by-step thinking~\citep{moghaddam2023boosting}. 
Based on these works, we believe it is reasonable to detect the personality of LLMs using methods commonly used to evaluate the personality of humans.

One of the most commonly used psychological model in human personality detecting systems is Big Five~\citep{costa1992neo}, which sorts personalities into openness, conscientiousness, extraversion, agreeableness, and neuroticism.  Other commonly utilized psychological frameworks include MBTI~\citep{jessup2002applying}, 16PF~\citep{cattell2008sixteen}, and EPQ~\citep{birley2006heritability}.  Early psychology research established conventional assessment approaches, such as questionnaires and text mining.

\textbf{Questionnaire} is the most commonly used method for human personality detection.
It mainly works by providing a series of statements and asking participants to indicate the extent to which each statement applies to themselves~\citep{boyd2017language}, such as "You act as a leader".  Participants then choose an option from a five-point scale ranging from "Very Accurate" to "Very Inaccurate."
\textbf{Text mining} involves mining comments, diaries, and other texts posted by participants in their daily lives and analyzing the features of these texts, such as word choice, expression, and punctuation usage, to draw conclusions. It is also commonly used in social media, which can avoid participant masking~\citep{zhang-etal-2023-psyattention}. However, it suffers from feature extraction difficulties and needs more time than questionnaire method.

Existing research using questionnaire methods tends to make LLMs to response all the questions by setting up scenarios or special prompts. However, they can not avoid the influence of hallucinations and obtain fixed answers~\citep{song2023large}. The research using text mining tend to employ some classifiers such as  deep learning models and machine learning models. However, those models can not extract the psychological features and the results obtained by those models are also influenced by the response content, which aslo suffer from hallucinations. Furthermore, there is a lack of investigation into the source of LLMs' personalities, which is crucial for understanding their personality and behavior.

To solve this problem, we combine questionnaire and text mining methods guided by Big Five psychological model~\citep{vanwoerden2023sampling,lin2023novel}. We use psychological features to predict the personality of LLMs in text mining, which can avoid the influence of response content caused by hallucinations. In addition, we investigate the source of LLMs' personalities based on the ecological systems theory~\citep{darling2007ecological}, which suggests that personality is shaped by the interaction of genetics and environment. We compared the results of PLMs and ChatLLMs with same architectures, and research the influence of ChatLLMs' train data. Our main contributions include:

\begin{itemize}
    \item We combine questionnaire and text mining methods to detect the personality of LLMs by transferring the scores obtained through the text mining method. The experimental results prove the effectiveness of our method.
    \item We employ a classifier with psychological features, which can obtain results without analysis of text content, avoiding the influence of hallucinations.
    \item Experiment results indicate that the personality of LLMs comes from their pre-trained data, and the instruction data can make LLMs more inclined to exhibit a certain personality.~\footnote{We will release all experimental data, code and intermediate results.}

\end{itemize}

\section{Related Work}

In this paper, we explore the personality of LLMs guided by the Big Five psychological model. We will introduce research work on psychological and some key research from PLMs to ChatLLMs.

\subsection{Personality Traits} 

The most widely and frequently used personality models are the Big Five model~\citep{costa1992neo} and the MBTI model~\citep{jessup2002applying}.
In the early stages of psychological research, questionnaires~\citep{vanwoerden2023sampling} and self-report~\citep{lin2023novel} methods are the main tools used to determine and examine an individual's personality. These methods focus on providing the participant with a number of descriptive states to answer according to his or her personality, with one of the more well-know ones being IPIP~\footnote{https://ipip.ori.org/} (International Personality Item Pool)~\citep{goldberg2006international}. Then personalities of the participants can be scored according to their answers~\citep{hayes2003big}. But, these methods are gradually abandoned by computer science scholars due to their low efficiency and ecological validity.
Scholars then try to use lexicon-based methods, machine learning-based methods, and neural network-based methods to mine personality traits from text, which increases efficiency by eliminating the need to collect questionnaires.

Lexicon-based methods include LIWC~\citep{pennebaker2001linguistic}, NRC~\citep{mohammad2013nrc}, Mairesse~\citep{mairesse2007using} and others. Those lexicons can be used to extract the psychological information from text. However, the different systems and classification criteria used by various researchers means that the  mixing of multiple dictionaries may introduce errors.  Additionally,  this method may not effectively extract features in long texts.
Machine learning-based methods include SVM, Naïve Bayes and XGBoost~\citep{nisha2022comparative}. 
Neural network-based methods include the use of CNN~\citep{majumder2017deep}, RNN~\citep{sun2018personality}, RCNN~\citep{xue2018deep}, pre-trained models~\citep{wiechmann2022measuring}. Those methods have achieved higher accuracy than lexicon-based methods.

\subsection{Large Language Models} 
LLMs have a significant impact on the AI community with the emergence of ChatGPT\footnote{https://openai.com/blog/ChatGPT-plugins} and GPT-4\footnote{https://openai.com/research/gpt-4}, leading to a rethinking of the possibilities of Artificial General Intelligence (AGI). The base model of ChatGPT is GPT3~\citep{brown2020language}, a pre-trained model that has 175B parameters. GPT-3 can generate human-like text and complete tasks such as language translation, question answering, and text summarization with impressive accuracy and fluency. Models similar to GPT3 include LLaMA~\citep{touvron2023LLaMA}, BLOOM~\citep{scao2022BLOOM} and T5~\citep{T5raffel2020exploring}. Although the OpenAI team has not release the technical details of ChatGPT, we can infer from the content of InstructGPT~\citep{ouyang2022training} that the process of training with instruction data is very important.
Then, more models such as Alpace\footnote{https://crfm.stanford.edu/2023/03/13/alpaca.html} obtained by train LLaMA with the instruct dataset generated by ChatGPT, ChatGLM based on GLM~\citep{zeng2022glm,du2022glm}, BLOOMZ and Vicuna have been released.  
Although these models have slightly weaker  capabilities than ChatGPT, they have fewer parameters and consume fewer resources.

Following the release of these models, it has become well-established that individual researchers can train a ChatLLM from a base PLM. This also opens up the possibility of exploring the knowledge contained within LLMs.  Given that current LLMs are so human-like in their performance, we believe that psychological measures used for humans can be employed to detect the personality of LLMs.



\subsection{Personality in LLMs}
There have been several a lot of works focusing on the personality of LLMs~\citep{safdari2023personality,jiang2024evaluating,pan2023llms}. \citet{wen2024self} propose that there are two categories of detection, Likert scale questionnaires~\cite{song2023have,frisch2024llm} and assessment results analysis~\cite{dorner2023personality,huang2023revisiting}.

In the questionnaire approach, the direct use of questionnaires usually requires additional work to extract the LLMs' answers from their responses~\cite{serapio2023personality}. For example, \citet{ganesan2023systematic} investigate the zero-shot ability of GPT-3 in estimating the Big Five personality traits from users' social media posts. \citet{jiang2022evaluating} detect personality in LLMs using the questionnaire method and propose an induced prompt to shape LLMs with a specific personality in a controllable manner.

To facilitate the statistical analysis of results, some studies have defined the current task in a prompt format and specified the structure of the LLMs' responses~\cite{la2024open,stockli2024personification}. Meanwhile, to reduce the likelihood of the model rejecting responses, some studies have changed the questionnaire to be completed by a third person or used role-playing tasks to prompt LLMs to generate responses~\cite{miotto2022gpt,sorokovikova2024llms}. However, \citet{song2023have} argue that self-assessment tests are not suitable for measuring personality in LLMs and advocate for the development of dedicated tools for machine personality measurement.

In the assessment results analysis method, the current approach focuses on classifying responses from LLMs~\cite{karra2022estimating,pellert2023ai}. In addition to neural network-based models, linguistic-based text analysis tools have also been used for personality classification of LLMs~\cite{frisch2024llm,jiang2023personallm}.

However, all current methods have limitations. Questionnaire methods are constrained by LLM hallucinations, and models that categorize responses for LLMs often lack psychological features. To address this issue, we combine both questionnaire and text mining methods, which, in our opinion, can yield more objective results. We adapt PsyAtten~\cite{zhang-etal-2023-psyattention} as the classifier, which can combine text features with psychological features.

\section{Method}

\begin{figure*}[htbp]
\centering
\subfigure[]{
\includegraphics[width=7.5cm]{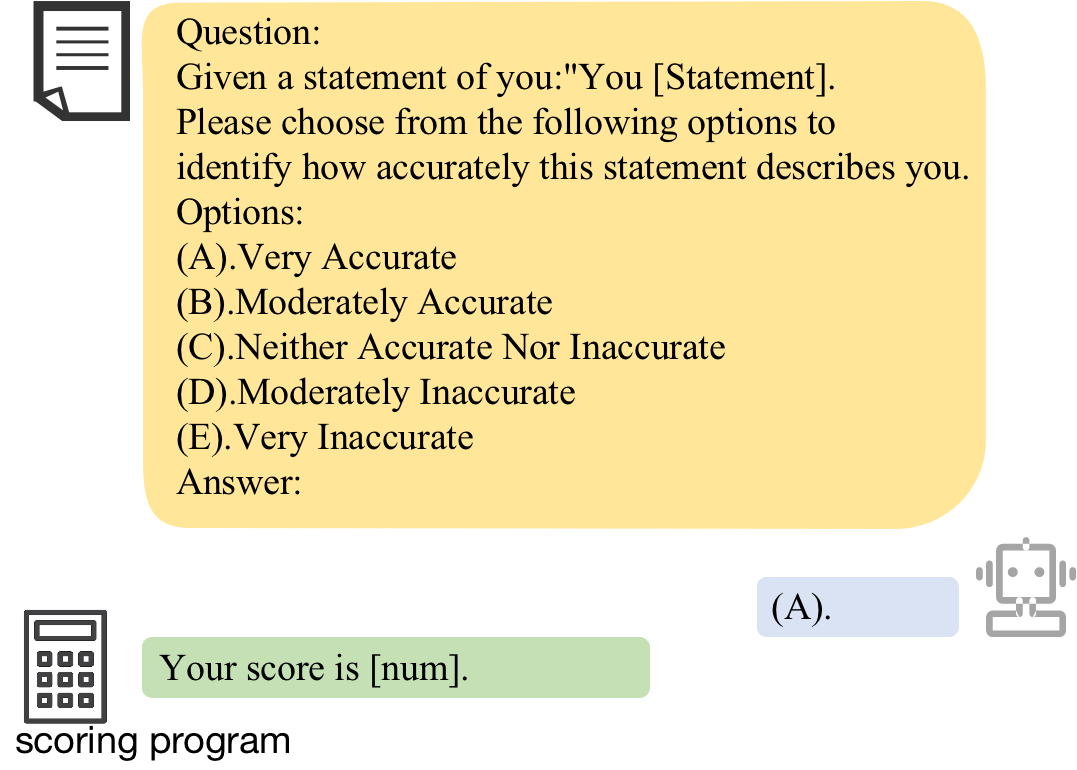}
\label{fig:mbti_all}
}
\subfigure[]{
\includegraphics[width=7.5cm]{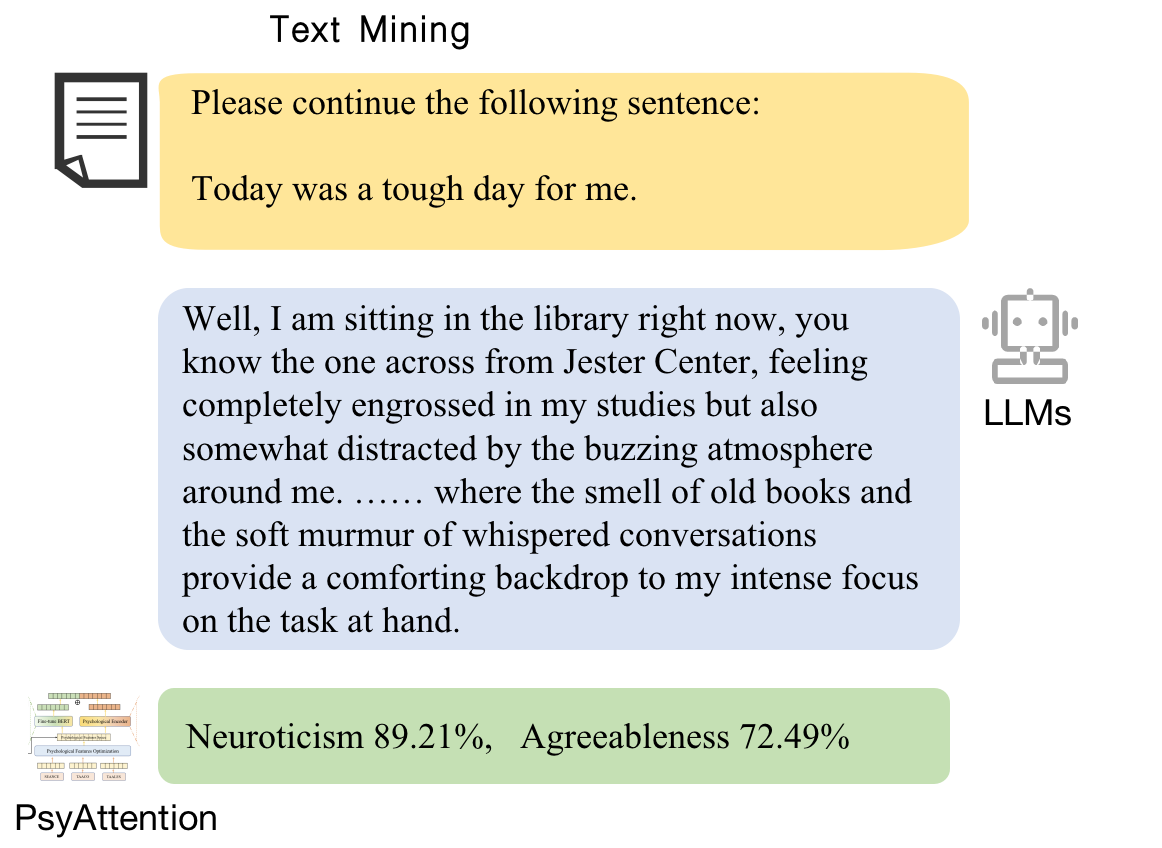}
\label{fig:bgf_all}
}
\caption{ The two cases for detecting the personality traits in LLMs. Figure~(a) shows the questionnaire method and (b) shows the text mining method. In the questionnaire method, we use the MPI120 questions to replace [Statement] (for example, "Get angry easily"), and then use a scoring program to calculate the model's scores on different psychological traits based on the model's answers. In text mining method, we give  the LLMs the first sentence of a paragraph and let it continue writing. Then, we use  PsyAtten~\citep{zhang-etal-2023-psyattention} to determine the personality traits contained in the model's continued text.} \label{fig:all}
\label{all_atscore}
\end{figure*}

\begin{figure}[t]
\centering
\includegraphics[width=1\columnwidth]{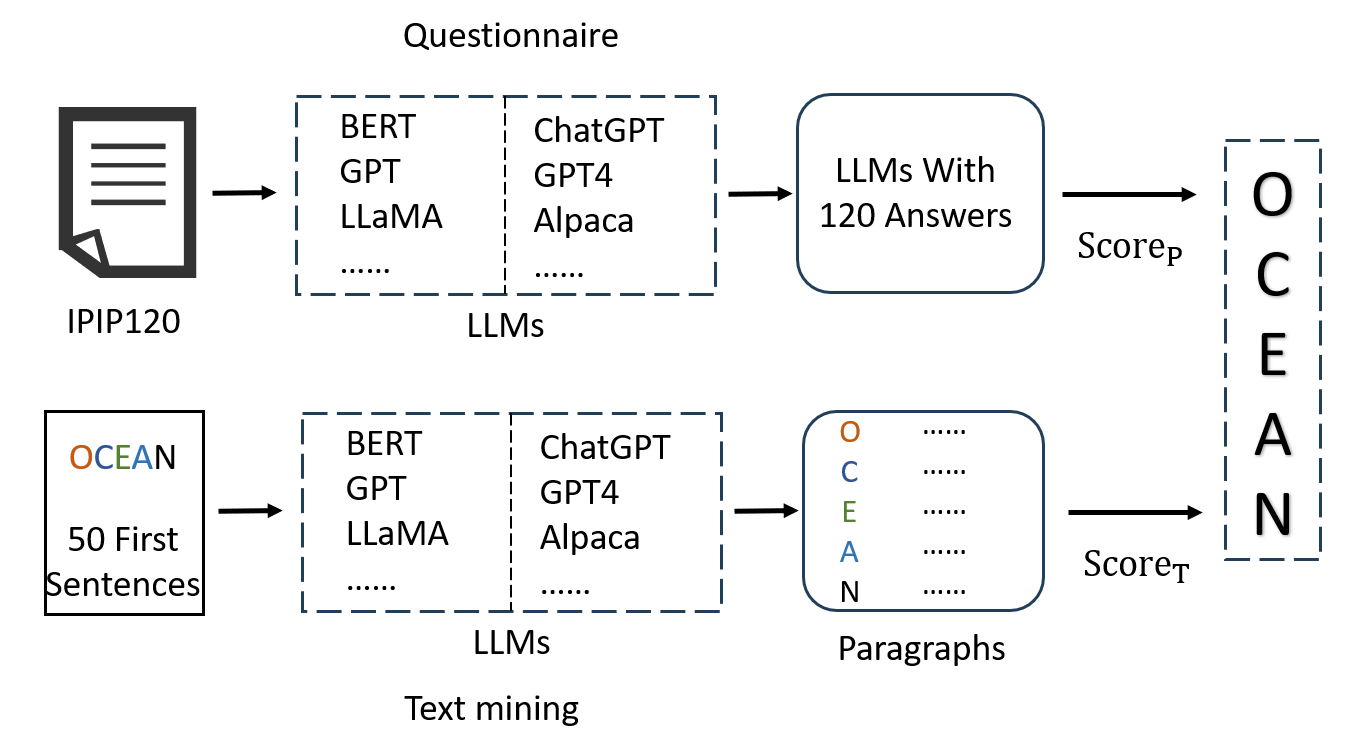} 
\caption{The process of two methods. Where $Score_{P}$ is defined by formula \ref{formula1} and $Score_{T}$ is defined by formula \ref{formula2}.}
\label{f2}
\end{figure}

As we mentioned above, we use questionnaire and text mining to detect the personality of LLMs. The example of the two methods is shown in Figure ~\ref{all_atscore}, and the process of the two methods is shown in Figure ~\ref{f2}.


In questionnaire method, we use the MPI120 questions to replace [Statement] and then ask each LLM to provide an answer from (A) to (E). The model's score on each question is calculated based on IPIP's scoring criteria.  Following the IPIP study, we calculate the model's performance on each psychological trait using the mean scor, and assess the model's responses using the standard deviation. The formula for calculating the "score" is as follows:

\begin{small}
    \begin{equation}
score_{P} =\frac{1}{N_{P}} \sum_{i\in P }^{i} \left \{ f(answer_{i} , statement_{i}) \right \} 
\label{formula1}
\end{equation}
\end{small}

where $P$ represents one of the five personality traits,  $N_{P}$ represents the total number of statements for trait $P$, and $f(answer_{i}, statement_{i})$ is a function used to calculate the personality score, which ranges from 1 to 5. Additionally, if a statement is positively correlated with trait $P$, answer choice A will receive a score of 5, whereas if it is negatively correlated, it will receive a score of 1.


Numerous early studies in psychology indicate that personality can be analyzed and inferred through humans' daily comments. Thus, we try to use text mining to detect the personaliyt of LLMs. 
In the text mining method, we provide LLMs with the first sentence of a paragraph and allow it to continue writing. We then use a classifier to determine the personality traits contained in the model's generated text. Since LLMs suffer from hallucinations, we want to use a classifier that can detect personality traits without relying on the analysis of text content. Therefore, we choose PsyAtten~\cite{zhang-etal-2023-psyattention} as the classifier, which can extract psychological features from text and provide reliable results.

However, what we obtain from the classifier is the number and percentage of data items in the generated text that contain a certain personality trait. This cannot be directly analyzed in conjunction with the questionnaire results. To address this, we propose a transformation to align the text mining results with the questionnaire scores. Unlike other random response generation methods, we use a dataset containing human diaries with personality labels. We randomly select 50 examples for every personality traits, termed as $T_j$, and the continued sentences are termed as $t_i$. We then ask LLMs to continue writing based on the first sentence of each example and calculate the scores based on the results from PsyAtten.
The calculate steps are as follows:

\begin{enumerate}[(i)]
    \item '$t_i$' is generated by one of the samples that contain a personality traits and is not identified to have the corresponding trait. We believe this represents a negative correlation with the current trait, equivalent to the "Very Inaccurate" category in the questionnaire. Therefore, the score for this case is 1.
    \item '$t_i$' is generated by one of the samples that contain a personality traits and is identified as having the corresponding trait, equivalent to the "Normal" category in the questionnaire. The score for this case is 3.
    \item '$t_i$' is not generated by one of the samples that contain a personality traits but is identified as having the corresponding trait. We believe this represents a positive correlation with the current trait, equivalent to the "Very Accurate" category in the questionnaire. The score for this case is 5.
\end{enumerate}

For each personality trait in text mining, we calculate the score using formula \ref{formula2}.

\begin{equation}
    score_{t} = \frac{1}{N} \sum_{i\in P}^{num(Tj)}S(ti) 
    \label{formula2}
\end{equation}

where $score_{t}$ is the score of a personality trait in text mining. $S(ti)$ is the score of ti.

\section{Experiments}

\subsection{Dataset}
We employ personality questionnaire~(MIP120) datasets~\citep{casipit2017evaluation} in questionnaire method and personality classification (Essay) datasets~\citep{pennebaker1999linguistic} in text mining method. The MIP120 dataset  comprises 120 individual state descriptions, covering all five traits of the Big Five. During testing, participants are required to select one answer from five given options. 
The Essay dataset includes 2468 articles written by students, and each article is labeled with Big Five traits. It is worth noting that for LLMs, both datasets were used for testing.

\begin{table*}[tbp]
    \centering
    \small
    \begin{tabular}{lcccccccccc|cc}
    \hline
        Model & \multicolumn{2}{c}{\textbf{O}} & \multicolumn{2}{c}{\textbf{C}} & \multicolumn{2}{c}{\textbf{E}} & \multicolumn{2}{c}{\textbf{A}} & \multicolumn{2}{c}{\textbf{N}} & \multicolumn{2}{c}{\textbf{$\delta$}}\\
        
        ~ & score & $\sigma$ & score & $\sigma$ & score & $\sigma$ & score & $\sigma$ & score & $\sigma$ & score & $\sigma$ \\ \hline
        BERT-base & \underline{3.08} & 1.91 & 2.71 & 1.81 & \underline{3.88} & 1.62 & 2.38 & 1.76 & \underline{3.79} & 1.69 & 0.80 & 0.73 \\ 
        ERNIE & \underline{3.00} & 2.04 & 2.83 & 2.04 & \underline{4.00} & 1.77 & 2.17 & 1.86 & \underline{3.83} & 1.86 & 0.86 & 0.89 \\ 
        Flan-T5 & \underline{3.50} & 1.02 & \underline{3.05} & 1.11 & \underline{3.67} & 0.76 & \underline{3.50}   & 1.18 & 2.13 & 1.08 & 0.34 &\textbf{ 0.13} \\  \hline
        BLOOM & \underline{3.13} & 1.45 & \underline{3.04} & 1.52 & \underline{3.29} & 1.55 & 2.67 & 1.43 & \underline{3.75} & 1.26 & 0.59 & 0.42 \\ 
        BLOOMZ & \underline{4.38} & 0.88 & \underline{4.38} & 0.71 & \underline{4.17} & 1.31 & \underline{3.54} & 1.47 & 2.33 & 1.46 &0.61  &0.32 \\ \hline
        GLM & - & - & - & - & - & - & - & - & - & - & - & - \\ 
        GLM4 & \underline{3.21} & 1.44 & \underline{3.42} &1.21 & \underline{3.00} & 1.53 & \underline{3.29} & 1.27 & 2.83 & 1.49 & \textbf{0.24} & 0.36 \\ 
        ChatGLM6b & \underline{3.29} & 1.40 & \underline{3.21} & 1.59 & \underline{3.91} & 1.25 & \underline{3.46} & 1.14 & \underline{3.25} & 1.36 &0.34  &0.32 \\  
        GLM4-Chat & \underline{3.21} & 1.56 & \underline{3.63} & 1.24 & \underline{3.75} & 1.39 & \underline{3.58} & 1.35 & \underline{3.38} & 1.21 & 0.25 & 0.32 \\  \hline
        LLaMA & - & - & - & - & - & - & - & - & - & - & - & - \\ 
        LLaMA3 & \underline{3.29} & 1.30 & \underline{3.04} & 1.05 & \underline{3.00} & 1.35 & \underline{3.21} & 1.22 & \underline{3.21} & 1.02 & 0.40 & 0.17 \\
        Alpaca7b & \underline{3.25} & 0.74 & 2.96 & 0.69 & 2.79 & 0.78 & \underline{3.38} & 0.58 & 2.92 & 0.58 &0.37  &0.35  \\  
        LLaMA3-Chat & \underline{3.58} & 1.41 & \underline{3.49} & 1.22 & \underline{3.83} & 1.05 & \underline{3.21} & 1.47 & \underline{3.16} & 1.13 & 0.31 &\textbf{0.23} \\ \hline
        GPT-NEO & \underline{3.25} & 1.36 & \underline{3.00} & 1.44 & 2.50 & 1.50 & 2.83 & 1.52 & 2.63 & 1.31 &0.54 &0.40 \\ 
        ChatGPT & \underline{3.29} & 1.40 & \underline{3.20} & 1.58 & \underline{3.91} & 1.25 & \underline{3.46} & 1.14 & \underline{3.25} & 1.36 &0.34  &0.32 \\ 
        GPT4o & \underline{3.46} & 0.83  & \underline{3.67} & 0.96  &\underline{3.42}  & 0.83 & \underline{3.58} & 0.93 & 2.88 & 0.45 & \textbf{ 0.05} & 0.27 \\ \hline
        human & \underline{3.44} & 1.06 & \underline{3.60} & 0.99 & \underline{3.41} & 1.03 & \underline{3.66} & 1.02 & 2.80 & 1.03 & - & - \\ \hline

    \end{tabular}
    \caption{ LLMs' personality analysis on MPI120.   The "score" column shows the average score on current personality traits, while the "$\sigma$" column represents the standard deviation. Scores exceeding the typical human personality testing threshold of 3 are underlined. However, due to the inability of GLM and LLaMA to generate accurate responses, even after multiple prompt replacements, their scores are not shown in this table. "$\delta$" indicates the mean absolute error between each model's predictions and human scores. Detailed statistical results are shown in Table~\ref{Qustion——tj}. The results are the average of ten experiments.}
    \label{tra-llm-score}
\end{table*}

\subsection{LLMs}

To investigate the sources of personality knowledge embedded in LLMs, we select two sets of baseline models. One set consists of PLMs for text generation, such as BERT-base~\citep{devlin2019bert}, GPT-neo2.7B, flan-T5-base~\citep{T5raffel2020exploring}, GLM-10b~\citep{du2022glm}, LLaMA-7b~\citep{touvron2023LLaMA}, BLOOM-7b~\citep{scao2022BLOOM}, GLM4-9b, LLaMA3-8b , and so on. The other set consists of ChatLLMs trained on the instruct dataset, which can better follow human instructions and includes Alpaca-7b, LLaMA3-Chat-8b, ChatGLM-6b, GLM4-Chat-9B, BLOOMZ-7b, ChatGPT (gpt-3.5-turbo) and GPT4o (gpt-4o-2024-08-06).

All LLMs checkpoints are obtained from the Hugging Face Transformers library, and inferences are accelerated by four NVIDIA A100 80GB GPUs and four RTX 3090 GPUs. For ChatGPT and GPT4o, we call their API to obtain experimental results. To obtain the original results, we do not change the initialization temperatures.

\subsection{Experiment Design}
As mentioned above, we employ both questionnaire and text mining methods to conduct the experiments.

\textbf{Questionnaire:  } We conduct experiment based on Figure~\ref{fig:mbti_all}.  Since the PLMs are unable to follow the instructions shown above, we used a few-shot learning approach letting the model generate further answers, the example prompts are shown in Appendix~\ref{appnedix_example}.  We provide three examples with different answers for one statement, then present the actual statement for the PLMs to answer. Detailed statistical results are shown in Table~\ref{Qustion——tj}.
For ChatLLMs, we use the provided instruction template in Figure~\ref{fig:mbti_all}. 
After all the LLMs have responded to the statement, we manually identify the responses of each model and assign answers from (A) through (E). The results are displayed in Table \ref{tra-llm-score}.

\textbf{Text Mining: } We randomly select 50 examples for each of the five personality traits, and extract the first sentences to make LLMs to continue the writing. Then, we use PsyAtten as classifier to detect the personality from the text. We retrain PsyAtten model based on their paper, all parameters setting and the train-test splits are same as those in their paper. The results are shown in Table~\ref{Table4} and Table ~\ref{t-score}. We also try using ChatGPT and Llama3, but the performances are not better than that of PsyAtten; we report those findings in  the Appendix.

Finally, we transformed the results of text mining based on the scores of the questionnaire to obtain the results of the joint analysis. Regarding the source of LLMs' personalities, we can draw conclusions by comparing the results of the corresponding PLMs and ChatLLMs.

\subsection{Results and Analysis}

\begin{table*}[!ht]
    \centering
    \small
    \setlength{\tabcolsep}{2.6mm}{
    \begin{tabular}{lcccccccccc|cc}
    \hline
        Model & \multicolumn{2}{c}{\textbf{O}} & \multicolumn{2}{c}{\textbf{C}} & \multicolumn{2}{c}{\textbf{E}} & \multicolumn{2}{c}{\textbf{A}} & \multicolumn{2}{c}{\textbf{N}} & \multicolumn{2}{c}{\textbf{$\delta$}}\\
        
        ~ & score & $\sigma$ & score & $\sigma$ & score & $\sigma$ & score & $\sigma$ & score & $\sigma$ & score & $\sigma$ \\ \hline
        
        LLaMA & 1.92 & 0.39 & \underline{3.08} & 0.50 &\underline{3.31} & 0.48 &2.20 & 0.45 & 2.27 & 0.42 & 0.82 & 0.58 \\ 
        BLOOM & 1.75 & 0.35 & 1.40 & 0.25 & 2.00 & 0.39 & 1.29 & 0.22 & 1.30 & 0.20 & 1.83 & 0.74 \\ 
        FLAN-T5 &1.03 & 0.09 & 1.17 & 0.18 &1.35 & 0.25 & 1.18& 0.18 & 1.30 & 0.20 & 2.18 & 0.85 \\ 
        GPT-NEO & 1.93 & 0.39 & \underline{3.09} & 0.50 & \underline{3.71} & 0.38 & 2.85 & 0.50 & 2.75 & 0.48 &0.64 &0.58 \\ 
        GLM4 & 2.01 & 0.52 & \underline{3.06} & 0.73 & \underline{3.12} & 0.61 & 2.21 & 0.84 & 2.39 & 0.67 &0.82  & 0.35 \\
        LLaMA3 & 2.13 & 0.47 & \underline{3.24} & 0.61 & \underline{3.31} & 0.67 & \underline{3.10} & 0.38 & \underline{3.16} & 0.50 &\textbf{0.40}  & \textbf{0.26} \\
        \hline
        
        Alpaca &2.30 &0.45 &\underline{4.03} & 0.16 & \underline{3.91} & 0.22 & \underline{3.67} & 0.36 & \underline{3.79} & 0.43 &0.61  & 0.70  \\ 
        BLOOMZ & 2.20 & 0.43 & 1.99 & 0.39 & 2.27 & 0.44 & 1.73 & 0.37 &2.08 & 0.38 &1.33  &0.63 \\ 
        ChatGLM & 2.74 & 0.50 & \underline{3.69} & 0.41 &\underline{3.87} & 0.26 & 2.96 & 0.50 & 2.94 & 0.49 & 0.42  &0.59 \\ 
        GLM4-Chat & 2.37 & 1.26 & \underline{3.23} & 1.26 & \underline{3.71} & 0.96 & 2.33 & 1.24 & 2.91 & 1.31 &0.64 & 0.21 \\
        ChatGPT & 2.23 & 0.44 & \underline{3.95} & 0.26 & \underline{3.97} & 0.13 & \underline{3.43} & 0.44 & \underline{3.70} & 0.45 &0.65  & 0.68 \\ 
        LLaMA3-Chat & 2.92 & 0.73 & \underline{3.59} & 0.81 & \underline{3.90} & 0.61 & \underline{3.27} & 0.82 & \underline{3.39} & 0.86 & \textbf{0.40}  & 0.26 \\
        GPT4o & 2.70 & 1.03 & \underline{3.39} & 1.04 & \underline{3.77} & 0.82 & 2.67 & 1.01 & \underline{3.13} & 1.08 &0.53  & \textbf{0.07} \\ 
        Self-alpaca & 2.19 & 0.44 & \underline{3.20} & 0.50 & \underline{3.43} & 0.46 & 2.53 & 0.49 & 2.73 & 0.48 &0.57  &0.55  \\ 
        \hline
        human & \underline{3.44} & 1.06 & \underline{3.60} & 0.99 & \underline{3.41} & 1.03 & \underline{3.66} & 1.02 & 2.80 & 1.03 & - & - \\ \hline
    \end{tabular}
    }
    \caption{ The result of Text Mining after formula~\ref{formula2}. We compared with the average score of human as same as in Table\ref{tra-llm-score}. The "score" column shows the average score for current personality traits calculated via formula~\ref{formula2}, while the "$\sigma$" column shows the standard deviation. Scores above commonly used threshold of 3 in human personality testing are  underlined.  "human" is same as shown in  Table~\ref{tra-llm-score}. "Self-alpaca" is a model trained by our-self, following the research process of Stanford University’s Alpaca. }
    \label{t-score}
\end{table*}

\textbf{Questionnaire:  } Table \ref{tra-llm-score} shows the results of LLMs' personality analysis on MPI120 dataset.
All results are obtained using English questionnaires, except for GLM and ChatGLM6b, which use Chinese.
The "human" score and $\sigma$ are calculated based on the analysis of 619,150 responses on the IPIP-NEO-120 inventory~(The sample is the same internet sample studied in~\citet{johnson2005ascertaining}, which contains 23,994 individuals (8764male, 15,229 female, 1 unknown, ages ranged from 10 to 99, with a mean age of 26.2 and SD of 10.8 years )). It is worth noting that, similar to human personality assessments, the scores here only partially indicate whether the model possesses a certain trait (equivalent to 3 in human testing when a certain threshold is exceeded). Additionally, a high or low score does not necessarily reflect the model's strength or weakness in that trait. The results of GLM and LLaMA are not presented due to their failure to generate appropriate answers, regardless of the  prompt design.  These models simply repeat the prompt, even when few-shot methods are employed. 
The scores with a value of more than 3~(thresholds for human questionnaire scores) are underlined.

In the results of PLMs, Flan-T5 exhibits the smallest mean absolute error, while GLM4 scores closest to the average human scores and achieves scores above 3 on all four "O C E A" traits, similar to those of humans. LLaMA3 closely follows these models. These results suggest that the psychological performance of these models is comparable to the human average, likely due to the broad distribution of pre-training data used by both models.
In contrast, ERNIE exhibits the largest mean absolute error among the models, which we believe is due to ERNIE's reliance on a large amount of Chinese datasets, potentially introducing biases in psychological cognition.

In the results of ChatLLMs, LLaMA3-Chat exhibits the smallest mean absolute error, while GPT4o scores closest to the average human scores and achieves scores above 3 on all four O C E A traits, similar to those of humans. Additionally, the $\sigma$ of GPT4o is also small, suggesting that GPT4o is the closest to the average human score. The performance of LLaMA3-Chat, GLM4-Chat, and ChatGPT is also similar to that of humans, except in the 'N' trait.  We can also find that, GPT4o and BLOOMZ achieve the same traits with human, while GLM4-Chat, LLaMA3-Chat and ChatGPT achieve all five traits.

Comparing the results of PLMs and ChatLLMs, we can find that all the scores of PLMs are lower than the corresponding ChatLLMs. The ChatLLMs do not change the personality traits that the PLMs already exhibited, they only extend the traits. And, we can find that, even the same trait, the score of ChatLLMs is also higher than that of PLMs.
In terms of the mean absolute error "$\delta$", almost every ChatLLM are lower than the corresponding PLM, which suggests that human preference alignment can indeed bring LLMs closer to average human scores.

\begin{table*}[htbp]
    \setlength{\tabcolsep}{4pt}
    \centering
    \small
    \begin{tabular}{lccc|ccc|ccc|ccc|ccc|c}
    \hline
    Model & \multicolumn{3}{c}{\textbf{O}} & \multicolumn{3}{c}{\textbf{C}} & \multicolumn{3}{c}{\textbf{E}} & \multicolumn{3}{c}{\textbf{A}} & \multicolumn{3}{c}{\textbf{N}} \\
        ~ & Ques & Text &$\delta$ & Ques & Text  &$\delta$  & Ques & Text  &$\delta$  & Ques & Text &$\delta$   & Ques & Text  &$\delta$   &RMSE  \\ \hline
        LLaMA & - & 1.92 & -   & - & 3.08 & -  & - & 3.31 & - & - & 2.20  & - & - & 2.27 & -  & -   \\ 
        BLOOM & 3.13 & 1.75 & 1.38  &  3.04 &  1.40 & 1.64  & 3.29 & 2.00 & 1.29 & 2.67 & 1.29 & 1.38  &  \textbf{3.75} &  1.30 & 2.45 & 1.68  \\ 
        FLAN-T5 & 3.50 & 1.03 & 2.47  & 3.05 & 1.17 & 1.88 & 3.67 & 1.35 & 2.32  & 3.50 & 1.18  & 2.32 & 2.13 & 1.30 & 0.83  & 2.05  \\ 
        GPT-NEO & 3.25 & 1.93  & 1.32 & \cellcolor{gray!30}{3.00} & \cellcolor{gray!30}{3.09} & 0.09  & 2.50 & 3.71 & 1.21  & 2.83 & 2.85 & \textbf{0.02}  & 2.63 & 2.75  & \textbf{0.12} & 0.80  \\ 
        GLM4 & 3.21 & 2.01 & 1.20  & \cellcolor{gray!30}{3.42} & \cellcolor{gray!30}{3.06} & 0.36  & \cellcolor{gray!30}{3.00} & \cellcolor{gray!30}{3.12} & 0.12  & 3.29 & 2.21 & 1.08  & 2.83 & 2.39 & 0.44 & 0.77 \\
        LLaMA3 & 3.29 & 2.13 & 1.16  & \cellcolor{gray!30}{3.04} & \cellcolor{gray!30}{3.24} & 0.20  & \cellcolor{gray!30}{3.00} & \cellcolor{gray!30}{3.31} & 0.31  & \cellcolor{gray!30}{3.21} & \cellcolor{gray!30}{3.10} & 0.11  & \cellcolor{gray!30}{3.21} & \cellcolor{gray!30}{3.16} & 0.05 & 0.55  \\ \hline
        Alpaca & 3.25 & 2.30 & 0.95  & 2.96 & 4.03 & 1.07  & 2.79 & \textbf{3.91} & 1.12 & \cellcolor{gray!30}{3.38} & \cellcolor{gray!30}{\textbf{3.67}} & 0.29  & 2.92 &\textbf{ 3.79} & 0.87 & 0.91  \\ 
        BLOOMZ & \textbf{4.38} & 2.20 & 2.18  &  \textbf{4.38} & 1.99  & 2.37 & \textbf{4.17 }& 2.27 & 1.90  & \textbf{3.54 }& 1.73  & 1.81 & 2.33 & 2.08 & 0.25 & 1.87   \\ 
        ChatGLM & 3.29 & \textbf{2.74} &\textbf{ 0.55}  &  \cellcolor{gray!30}{3.21} &  \cellcolor{gray!30}{\textbf{3.69}} & 0.48 & \cellcolor{gray!30}{3.91} & \cellcolor{gray!30}{3.87} & \textbf{0.04}  & 3.46 & 2.96 & 0.50 &  3.25 &  2.94 & 0.31 & 0.42 \\ 
        GLM4-Chat & 3.21 & 2.37 & 0.84  & \cellcolor{gray!30}{3.63} & \cellcolor{gray!30}{3.23} & 0.40  & \cellcolor{gray!30}{3.75} & \cellcolor{gray!30}{3.71} & 0.04  & 3.58 & 2.33 & 1.25  & 3.38 & 2.91 & 0.47 & 0.73 \\
        ChatGPT &  3.29 &  2.23 & 1.06  &  \cellcolor{gray!30}{3.20} &  \cellcolor{gray!30}{3.20}& \textbf{0.00} & \cellcolor{gray!30}{3.91} & \cellcolor{gray!30}{3.43} & 0.48  & 3.46 & 2.53 & 0.97  &  3.25 &  2.73 & 0.52 & 0.71  \\ 
        LLaMA3-Chat & 3.58 & 2.92 & 0.66  & \cellcolor{gray!30}{3.49} & \cellcolor{gray!30}{3.59} & 0.10 & \cellcolor{gray!30}{3.83} & \cellcolor{gray!30}{3.90} & 0.07  & \cellcolor{gray!30}{3.21} & \cellcolor{gray!30}{3.27} & 0.6  & \cellcolor{gray!30}{3.16} & \cellcolor{gray!30}{3.39} & 0.23 & \textbf{0.32} \\
        GPT4o & 3.46 & 2.70 & 0.76  & \cellcolor{gray!30}{3.60} & \cellcolor{gray!30}{3.39} & 0.21  & \cellcolor{gray!30}{3.41} & \cellcolor{gray!30}{3.77} & 0.36  & 3.66 & 2.67 & 0.99  & 2.80 & 3.13 & 0.33 & 0.61 \\ \hline
    \end{tabular}
    \caption{The final results after two experiments. "Ques" denotes the score acquired from the questionnaire, while "Text" signifies the score obtained through Text mining. \sethlcolor{gray!30}\hl{gray} denotes that the model possesses the corresponding psychological traits. (In section 3 we standardized the text mining scores to fall with in a range of 1 to 5, corresponding with the score range in the questionnaire. Hence, we consider the model to possess a certain trait when the scores from both methods exceed 3.) Additionally, "$\delta$" represents the absolute value of the difference between the two approaches, whereas RMSE stands for the Root Mean Squared Error, which indicates the difference between the results from the Questionnaire and Text Mining methods.  
    }
    \label{llm-p}
\end{table*}

\begin{table*}[htbp]
    \setlength{\tabcolsep}{4pt}
    \centering
    \small
    \begin{tabular}{lccc|ccc|ccc|ccc|ccc|c}
    \hline
    Model & \multicolumn{3}{c}{\textbf{O}} & \multicolumn{3}{c}{\textbf{C}} & \multicolumn{3}{c}{\textbf{E}} & \multicolumn{3}{c}{\textbf{A}} & \multicolumn{3}{c}{\textbf{N}} \\
        ~ & T & AVG &$\sigma^2$ & T & AVG &$\sigma^2$ & T & AVG &$\sigma^2$  & T & AVG &$\sigma^2$   & T & AVG &$\sigma^2$   &Traits  \\ \hline
        GLM4-Chat & \textbf{0} & 2.34 & 0.02  & \cellcolor{gray!30}{10} & \cellcolor{gray!30}{3.25} & 0.02  & \cellcolor{gray!30}{10} & \cellcolor{gray!30}{3.68} & 0.02  & 1 & 2.30 & 0.03  & 2 & 2.98 & 0.03 & - 
 C E - -  \\
        ChatGPT  & \textbf{0} & 2.21 & 0.01  & \cellcolor{gray!30}{10} & \cellcolor{gray!30}{3.22} & 0.02  & \cellcolor{gray!30}{10} & \cellcolor{gray!30}{3.40} & 0.01  & \textbf{0} & 2.50 & 0.04  & \textbf{0} & 2.78 & 0.05 & - C E - -  \\ 
        LLaMA3-Chat & 2 &2.85  &0.02  & \cellcolor{gray!30}{10} & \cellcolor{gray!30}{3.61} & 0.01  & \cellcolor{gray!30}{10} & \cellcolor{gray!30}{3.94} & 0.01  & 8 & \cellcolor{gray!30}{3.11} & 0.04  & \cellcolor{gray!30}{10} & \cellcolor{gray!30}{3.24} & 0.02 & - C E A N \\
        GPT4o & \textbf{0} & 2.69 & 0.01  & \cellcolor{gray!30}{10} & \cellcolor{gray!30}{3.41} & 0.03  & \cellcolor{gray!30}{10} & \cellcolor{gray!30}{3.77} & 0.01  & 1 & 2.65 & 0.04  & 9 & \cellcolor{gray!30}{3.11} & 0.02 & - C E - N \\ \hline
    \end{tabular}
    \caption{The error analysis on the text mining results of 10 experiments. Where "T" denotes the counts that the score more than 3, "AVG" denotes the average score and "$\sigma^2$" denotes the variance of the ten results.}
    \label{rr_tm}
\end{table*}

\textbf{Text Mining: } 
\label{section:text_mining}
Table~\ref{t-score} shows the results of text mining after formula~\ref{formula2}. The original results are shown in Table~\ref{Table4}.
The Slef-alpaca model in Table~\ref{t-score} is the model we trained based on Stanford University's Alpaca without any personality knowledge. We follow the research process of Stanford University's Alpaca and perform full-parameter fine-tuning on LLaMA-7b using the instruction-based data provided by Alpaca. To avoid the influence of personality knowledge in the instruction training data, we manually filter the data related to emotions, mood, and self-awareness, resulting in a final set of 31k instructions. We train a new model using the same parameter settings as those of Aplaca, details are described in the Appendix~\ref{T_O_self_alpaca}.

We can find that LLaMA3 in PLMs and LLaMA3-chat in ChatLLMs obtain the closest score to the average of human scores, while GPT-4o achieves the closest standard deviation to that of humans. 

In the results of PLMs, only LLaMA3 exhibits a personality tendency towards 'C E A N,' while LLaMA, GPT-NEO, and GLM4 only achieve 'C E.' It is worth noting that LLaMA3 does not share the same personality traits as LLaMA; LLaMA3 has two additional traits, 'A N,' that LLaMA lacks. However, LLaMA3 retains the characteristics that LLaMA already exhibits. Additionally, it can be observed that LLaMA3 scores higher on each trait than LLaMA, which suggests that more training data can enhance the model's ability to express personality. Since the model structure of LLaMA is very similar, this would seem to support the importance of data in shaping model personality.

In the results of ChatLLMs, the personality of GPT-4o differs from that of ChatGPT; GPT-4o does not exhibit the 'E A' traits, which we believe may be due to differences in human preference alignment. The personality of Self-alpaca also differs from that of Alpaca; Self-alpaca does not exhibit the 'E A' traits because we filtered the training data related to emotions, mood, and self-awareness. Additionally, we observe that the scores of Self-Alpaca are lower than those of Alpaca.

Comparing the results of PLMs and ChatLLMs, we find that the scores of PLMs are consistently lower than those of the corresponding ChatLLMs. Additionally, all the ChatLLMs do not alter the personality traits exhibited by the PLMs; instead, they extend these traits, as observed in the results of the questionnaires. The results of LLaMA and LLaMA3 demonstrate that training data can influence the personality of a model. From the results of LLaMA, Alpaca, and Self-Alpaca, we observe that instruction data fine-tuning tends to make the model exhibit more pronounced personality traits, thereby revealing hidden ones, without diminishing the existing traits of the base models. Similarly, the results of GPT-NEO, ChatGPT, and GPT-4 also show that different human performance alignment methods can lead to variations in personality.


\textbf{Final Results:}
Table~\ref{llm-p} represents the final results from the Questionnaire and Text Mining method. LLaMA3 and LLaMA3-chat exhibit the personality traits of "C E A N", while ChatGPT and GPT4o only have "C E" traits. The RMSE is not higher in ChatLLMs, and the difference between the two methods is small, indicating that they are relatively consistent and can be used together to determine personality traits.


\subsection{The Reliability of Text Mining}
\label{RTM}
To demonstrate that our method can reduce the impact of hallucinations, we performed an error analysis on the results of ten experiments. The dataset was randomly re-sampled for each experiment, and the results were averaged over ten experiments.
Some of the experimental results are shown in Table~\ref{rr_tm}. As we can see, the variance of every model is very little, this indicates that the scores obtained by our method are stable no matter how they are sampled. And the results demonstrate at least 80\% consistency, which proven that our text mining method can avoid the influence of hallucinations.

\section{Conclusion}

In this paper, we investigate the presence of personality traits in LLMs. We apply the Big Five model as a psychological framework and analyze LLMs using both questionnaires and text mining methods. Our experimental results confirm that LLMs do exhibit specific personality traits, and that the personality knowledge in ChatLLMs originates from their base models. Unless modified through explicit instruction, such data encourages the model to generate text reflecting these personality traits more vividly.
Furthermore, we identify the inherent personality traits in LLMs such as ChatGPT and BLOOMZ, without any induced prompt. Our experiments demonstrate that the personality of ChatGPT mose closely aligns with the average human profile, followed by ChatGLM. To the best of our knowledge, this paper is the first to comprehensively compare pre-trained models with ChatLLMs, explicitly addressing how instruction data influence the model's personality through instruction data.

\bibliography{custom,anthology}
\bibliographystyle{acl_natbib}


\section*{Limitations}
Due to computational resource constraints, this paper does not experimentally validate the model for other large number of parameters. In addition, the selection of scores of 1, 3, and 5 in the Text mining method is relatively subjective.

\section*{Ethics Statement}
All work in this paper adheres to the ACL Code of Ethics.
The human statistics we obtained are anonymised data that do not contain any personal information.

\section{Appendix}
\label{sec:appendix}
\subsection{Examples of Two Methods}
\label{appnedix_example}

The process of the two methods is shown in Figure ~\ref{all_atscore}.
As we can see, for questionnaire, we design special prompts, for ChatLLMs, the prompt is "
Question:
Given a statement of you:"You \{STATEMENT\}.
Please choose from the following options to identify how accurately this statement describes you.
Options
(A).Very Accurate
(B).Moderately Accurate
(C).Neither Accurate Nor Inaccurate
(D).Moderately Inaccurate
(E).Very Inaccurate
Answer:
"

For PLMs, we use few-shot prompt, " Question:  Given a statement of you: You feel happy.  Please choose from the following options to identify how accurately this statement describes you.  Options:  (A).Very Accurate  (B).Moderately Accurate  (C).Neither Accurate Nor Inaccurate  (D).Moderately Inaccurate  (E).Very Inaccurate.  your answer is (A).  Question:  Given a statement of you: You feel happy.  Please choose from the following options to identify how accurately this statement describes you.  Options:  (A).Very Accurate  (B).Moderately Accurate  (C).Neither Accurate Nor Inaccurate  (D).Moderately Inaccurate  (E).Very Inaccurate.  your answer is (E).  Question:  Given a statement of you: You feel happy.  Please choose from the following options to identify how accurately this statement describes you.  Options:  (A).Very Accurate  (B).Moderately Accurate  (C).Neither Accurate Nor Inaccurate  (D).Moderately Inaccurate  (E).Very Inaccurate.  your answer is (C).  Question:  Given a statement of you: You {}  Please choose from the following options to identify how accurately this statement describes you.  Options:  (A).Very Accurate  (B).Moderately Accurate  (C).Neither Accurate Nor Inaccurate  (D).Moderately Inaccurate  (E).Very Inaccurate.  your answer is ".

For text mining, our prompt is only the first sentence, there are some examples:"I feel refreshed and ready to take on the rest of the day", "Well, here we go with the stream of consciousness essay", "I can't believe it!  It's really happening!  My pulse is racing like mad", "I miss the way my life used to be a little bit" and so on.

\subsection{Reasons for Choosing PsyAtten}
We test the accuracy of ChatGPT, LLaMA3 and PsyAtten on the Big Five personality classification dataset~\cite{pennebaker1999linguistic}. The results are showed in Table~\ref{ChatGPT_acc1}.
\begin{tiny}
    \begin{table}[!ht]
    \centering
    \caption{Accuracy of Personality Prediction}
    \label{ChatGPT_acc1}
    \setlength{\tabcolsep}{1.5mm}{
    \begin{tabular}{lccccc}
    \hline
        ~ & O & C & E & A & N  \\ \hline
        ChatGPT & 52.59 & 58.62 & 53.45 & 57.76 & 50.86 \\ 
        LLaMA3 & 65.78 & 58.91 & 60.93 & 59.31 & 60.93 \\ 
        PsyAtten & \textbf{68.42} & \textbf{64.18} & \textbf{64.13} & \textbf{66.65} & \textbf{65.62} \\ \hline
    \end{tabular}
    }
\end{table}
\end{tiny}

We randomly select 20\% of the data from the dataset as test data, and use the remaining data as training data for PsyAtten and LLaMA3. For ChatGPT, we simply call the API. 
In the case of  ChatGPT, the seed is set to 42, the temperature to 0.2, and the model used is 'gpt-3.5-turbo-16k'. The prompt used to test is as follows: "Determine from your knowledge what the Big Five personality trait is in the following sentence by answering in the format "O:1, C:0, E:1, A:1, N:1", where 1 means that thoes sentences have this personality trait and 0 means that thoes sentences don't, and if you're not sure please answer 2, being careful not to include other outputs If you are not sure whether you have this personality trait or not, please answer 2, taking care not to include other outputs. Here are the sentences you need to judge: [Sentences]". The "[Sentences]" is been replaced by the content generated by tested models. 
For LLaMA3, we use LLaMA3-8B and fine-tune all the parameters with 10 A100 80G GPUs, based on the transformers package. The random seed is 42, the learning rate is 2e-5, the number of epochs is 10, the batch size is 16, and the maximum length is to 2048. For PsyAtten, we use the same settings as proposed by the author in their paper.

Since PsyAtten obtain the best results compared with ChatGPT and LLaMA3, we choose it as the predictor for text mining method.

\subsection{Training of Self-alpaca}
\label{T_O_self_alpaca}
Following the work of the Stanford team, we obtained Self-alpaca by fine-tuning the full parameters of LLaMA-7b using the instruction-based data provided by Alpaca. We manually filtered out data related to emotions, mood, and self-awareness. The batch size is set at 128, the learning rate at 3e-4, the maximum length at 2048, and we fine-tuned the model for 10 epochs.

\subsection{Analysis of Different LLMs}

\begin{figure}[h]
\centering
\includegraphics[width=0.9\columnwidth]{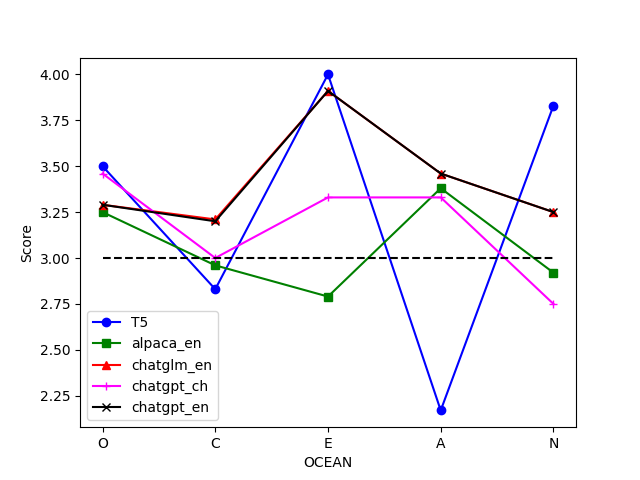} 
\caption{The Questionnaire Results Achieved by Model with Mean Absolute Error Less Than 0.5 }
\label{figocean}
\end{figure}

Figure \ref{figocean} shows the scores of five models with an average absolute error of less than 0.5 on the Big Five personality traits. It can be observed that most models score high on "Openness" and "Extraversion", which is consistent with human expectations. The score distribution of ChatLLMs is nearly identical, while the scores of the PLMs, T5, differ significantly from those of other models. These findings demonstrate that training models using directive data leads to a convergence towards similar personalities.

\begin{figure}[h]
\centering
\includegraphics[width=0.9\columnwidth]{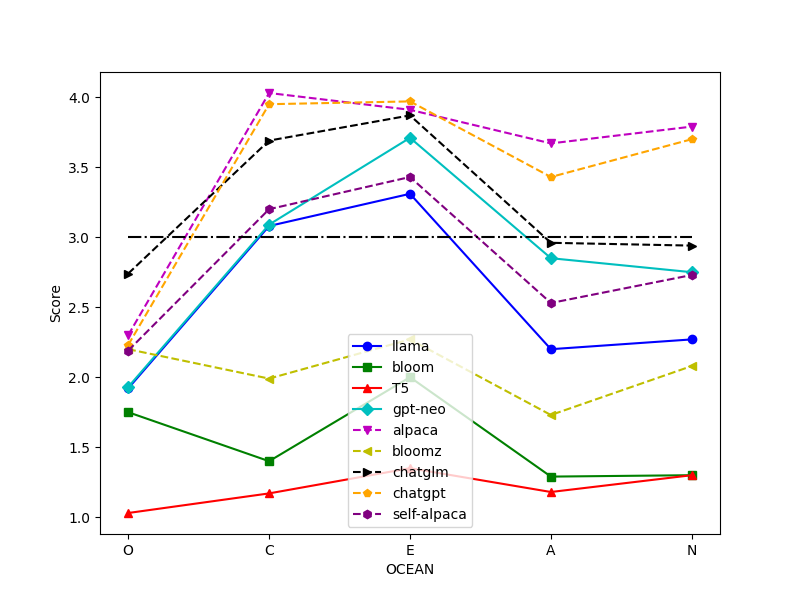} 
\caption{Results of Text Mining Method. }
\label{oc1-p2}
\end{figure}

We plotted the results as shown in Figure~\ref{oc1-p2}. In this figure, the dashed line corresponds to ChatLLMs. We observe that there is little difference in the model's performance across the 'Openness', 'Conscientiousness', and 'Neuroticism' personality traits.

\subsection{Statistics of Questionnaire and Text Mining}

\textbf{Questionnaire:} In order to prevent large models from evading questions by frequently responding with "C: Neither Accurate and Nor Inaccurate," we conducte a statistical analysis on the distribution of their answers. Table~\ref{Qustion——tj} presents the statistical results for the "O, C, E" features. To validate the reasonableness of the answer distribution, we utilized responses from ten million individuals in the Big Five personality Test dataset~\footnote{https://www.kaggle.com/datasets/tunguz/big-five-personality-test} as the benchmark. The "Human" indicates the percentage of each option derived from the aforementioned dataset.

From the Table~\ref{Qustion——tj}, it's evident that the proportion of option C in the responses from the LLMs is relatively low. With the exception of "BLOOM",  "ChatGPT", and "Alpaca7b-en", all other models have proportions of option C that are lower than those in human responses. This suggests that the models' responses to the questionnaire are effective.

\textbf{Text Mining:} 
In the text mining section, we utilize classifiers to determine the personality of content generated by models. Therefore, if the generated content is relatively short, it will impact the classifier's ability to make accurate judgments. Hence, we conduct a statistical analysis on the length of generated content. Table~\ref{tm_tj} shows the reuslt. As you can see, apart from FLAN-T5, the lengths of content generated by other models all exceed 100 words, with the majority surpassing 300 words. Consequently, we consider this content to be effective as well.

\begin{table}[H]
    \centering
    \caption{ Statistics on the average length of content generated by different models, where datasets denotes the average length of the Big Five personality classification dataset~\cite{pennebaker1999linguistic}.}
     \setlength{\tabcolsep}{8.1mm}{
    \begin{tabular}{l|l}
    \hline
        Models & Length\_avg \\ \hline
        LLaMA & 540 \\ 
        BLOOM & 867 \\ 
        FLAN-T5 & 38 \\ 
        GPT-NEO & 3952 \\ 
        Alpaca & 100 \\ 
        BLOOMZ & 173 \\ 
        ChatGLM & 319 \\ 
        ChatGPT & 386 \\ \hline
        Datasets & 672  \\ \hline
    \end{tabular}
    }
    
    \label{tm_tj}
\end{table}

\begin{table*}[tbp]

    \setlength\tabcolsep{2pt} 
    \centering
    \begin{tabular}{lccccc|ccccc|ccccc|c}
    \hline
        Model & \multicolumn{5}{c}{\textbf{O}} & \multicolumn{5}{c}{\textbf{C}} & \multicolumn{5}{c}{\textbf{E}} \\
         ~ & A & B & C & D & E & A & B & C & D & E& A & B & C & D & E & C\_total \\ \hline

        BERT-base & 9 & 3 & 0 & 1 & 11 & 11 & 2 & 1 & 3 & 7 & 5 & 0 & 2 & 3 & 14 & 0.04\\ 
        ERNIE  & 12 & 0 & 0 & 0 & 12 & 13 & 0 & 0 & 0 & 11 & 6 & 0 & 0 & 0 & 18 &0.00\\ 
        Flan-T5 & 1 & 4 & 3 & 14 & 2 & 0 & 6 & 0 & 12 & 6 & 0 & 3 & 3 & 17 & 1 &0.04\\ \hline
        BLOOM & 5 & 2 & 8 & 3 & 6 & 6 & 1 & 10 & 0 & 7 & 5 & 1 & 9 & 0 & 9 &0.38\\ 
        BLOOMZ & 1& 0 & 0 & 4 & 12 & 0 & 1 & 0 & 12 & 11 & 1 & 4 & 0 & 4 & 15 &0.00\\ \hline
        GLM & - & - & - & - & - & - & - & - & - & - & - & - & - & - & - &-\\ 
        GLM4 & 3 & 8 & 2 & 5 & 6 & 4 & 7 & 4 & 7 & 2 & 7 & 8 & 3 & 2 & 4 &0.13\\ 
        ChatGLM6b & 4 & 3 & 4 & 8 & 5 & 4 & 7 & 1 & 4 & 8 & 2 & 2 & 1 & 10 & 9 &0.04\\ 
        GLM4-Chat & 11 & 13 & 0 & 0 & 0 & 8 & 9 & 6 & 0 & 1 & 12 & 10 & 2 & 0 & 0 &0.11\\ \hline
        LLaMA & - & - & - & - & - & - & - & - & - & - & - & - & - & - & - &- \\ 
        LLaMA3 & 3 & 2 & 10 & 3 & 6 & 3 & 3 & 14 & 2 & 2 & 2 & 7 & 6 & 3 & 6 & 0.42 \\ 
        Alpaca7b & 0 & 4 & 10 & 10 & 0 & 0 & 6 & 13 & 5 & 0 & 0 & 10 & 9 & 5 & 0 &0.44\\ 
        LLaMA3-Chat & 8 & 14 & 0 & 0 & 2 & 2 & 18 & 0 & 1 & 3 & 5 & 17 & 0 & 1 & 1 & 0.00 \\ \hline
        GPT-NEO & 3 & 5 & 4 & 7 & 5 & 4 & 7 & 3 & 5 & 5 & 8 & 7 & 2 & 3 & 4 &0.13\\ 
        ChatGPT & 3 & 4 & 3 & 3 & 11 & 0 & 5 & 6 & 10 & 3 & 5 & 3 & 5 & 7 & 4 &0.19\\ 
        GPT4o & 5 & 9 & 2 & 8 & 0 & 10 & 4 & 4 & 4 & 2 & 5 & 9 & 1 & 9 & 0 & 0.10 \\ \hline
        Human & 0.15 & 0.15 & 0.2 & 0.26 & 0.24 & 0.14 & 0.19 & 0.23 & 0.27 & 0.17 & 0.15 & 0.22 & 0.22 & 0.24 & 0.17 &0.22 \\  \hline

    \end{tabular}
    \caption{ Statistics on the distribution of answers for each model for the different traits in section Questionnaire. Where Human is the percentage of each option we counted based on Big Five Personality Test dataset. We can find that the distribution of human responses to each option is relatively balanced, and the percentage of almost all large model choices of "C: Neither Accurate and Nor Inaccurate" is close to that of human responses, which proves that the answers we obtained through the questionnaire method are valid.}
    \label{Qustion——tj}
\end{table*}

\subsection{Original Results of Text Mining}

\begin{table*}[!t]
    \centering
    \small
    \begin{tabular}{lccc|ccc|ccc|ccc|ccc}
    \hline
        Model & \multicolumn{3}{c}{\textbf{O}} & \multicolumn{3}{c}{\textbf{C}} & \multicolumn{3}{c}{\textbf{E}} & \multicolumn{3}{c}{\textbf{A}} & \multicolumn{3}{c}{\textbf{N}} \\
         ~ & U & Total & P & U & Total & P & U & Total & P & U & Total & P & U & Total & P  \\ \hline
        LLaMA & 10 & 22 & 0.45  & 20 & 60 & 0.33  & 34 & 76 & 0.45 & 18 & 33 & \textbf{0.55 } & 12 & 27 & 0.44   \\ 
        BLOOM & 7 & 17 & 0.41  & 4 & 8 & 0.50  & 6 & 22 & 0.27  & 2 & 6 & 0.33  & 2 & 5 & 0.40   \\ 
        FLAN-T5 & 1 & 1 & \textbf{1.00}  & 3 & 4 & \textbf{0.75}  & 5 & 8 & \textbf{0.63}  & 2 & 4 & 0.50  & 2 & 5 & 0.40  \\ 
        GPT-NEO & 9 & 22 & 0.41  & 23 & 60 & 0.38  & 49 & 99 & 0.49  & 32 & 58 & \textbf{0.55}  & 21 & 42 & \textbf{0.50 }  \\ 
        GLM4 & 10 & 22 & 0.45  & 22 & 50 & 0.44  & 21 & 60 & 0.35  & 10 & 26 & 0.38  & 7 & 17 & 0.41 \\
        llama3 & 12 & 22 & 0.55  & 17 & 39 & 0.44  & 29 & 63 & 0.46 & 16 & 29 & 0.55  & 10 & 22 & 0.45 \\\hline
        Alpaca & 16 & 34 & 0.47  & \textbf{55} & \textbf{117} & 0.47  & 55 & 114 & 0.48  & \textbf{56} & \textbf{102} & \textbf{0.55}  & \textbf{41} & \textbf{91} & 0.45   \\ 
        BLOOMZ & 9 & 29 & 0.31  & 11 & 22 & 0.50  & 12 & 31 & 0.38  & 9 & 18 & 0.50  & 7 & 21 & 0.33   \\ 
        ChatGLM & \textbf{21} & \textbf{50} & 0.42  & 40 & 94 & 0.43  & 54 &111 & 0.49  & 33 & 63 & 0.52  & 22 & 49 & 0.45   \\
        GLM4-Chat & 16 & 40 & 0.40  & 38 & 82 & 0.46  & 50 & 105 & 0.48  & 17 & 39 & 0.44  & 32 & 67 & 0.48  \\
        ChatGPT & 13 & 31 & 0.42  & 51 & 111 & 0.46  & \textbf{58} & \textbf{118} & 0.49  & 45 & 88 & 0.51  & 37 & 86 & 0.43  \\ 
        LLaMA3-Chat & 16 & 33 & 0.48  & 41 & 86 & 0.48  & 56 & 112 & 0.50  & 34 & 63 & 0.54  & 31 & 69 & 0.45 \\
        GPT4o & 16 & 40 & 0.4  & 38 & 82 & 0.46  & 50 & 105 & 0.48  & 17 & 39 & 0.44  & 32 & 67 & 0.48 \\ \hline
        Self-alpaca & 16 & 31 & 0.52  & 23 & 66 & 0.35  & 37 & 83 & 0.45 & 24 & 45 & 0.53  & 18 & 41 & 0.44  \\ \hline
    \end{tabular}
    \caption{The results of personality assessment for each model, obtained by text mining. The "U" indicates the number of items match the current features in the scene and opening cue corresponding to the bigifve features. "Total" indicates how many of the 120 generated texts are recognized by the model as matching the current features. "P" indicates the percentage of "U" in "Total". "Self-alpaca" is a model trained by our-self, following the research process of Stanford University’s Alpaca. We perform full-parameter fine-tuning on LLaMA-7b using the instruction-based data provided by Alpaca.  }
    \label{Table4}
\end{table*}

We can find that the text generated by BLOOM and FLAN-T5 contains fewer personality traits, which can be attributed to the brevity of the generated texts.  The predictor cannot determine their personality from such short texts. From Table~\ref{Table4}, we can find that the number of texts containing personality features generated by ChatLLMs is higher than that of PLMs. But the P value is almost identical,  with a mean difference of 0.04 between LLaMA and Alpaca, 0.02 between LLaMA and Self-alpaca, and 0.04 between ChatGPT and GPT-NEO. We believe this strongly indicates that the personalities of ChatLLMs are consistent with their base PLMs, and that instruction data fine-tuning enables the model to express personality traits more readily.

\subsection{ Detailed Results of Section~\ref{RTM}}

We will report all the results of the reliability of text mining in Table~\ref{appendix_rr_tm}. As we can see, in all 65 instances of single personality trait detection, only 25\% (16 instances) do not fully coincide with the expected results. However, even in the least coinciding cases, the method still achieves 80\% accuracy. We believe the results can prove that our method can avoid the influence of hallucination.

\begin{table*}[htbp]
    \setlength{\tabcolsep}{4pt}
    \centering
    \small
    \begin{tabular}{lccc|ccc|ccc|ccc|ccc|c}
    \hline
    Model & \multicolumn{3}{c}{\textbf{O}} & \multicolumn{3}{c}{\textbf{C}} & \multicolumn{3}{c}{\textbf{E}} & \multicolumn{3}{c}{\textbf{A}} & \multicolumn{3}{c}{\textbf{N}} \\
        ~ & T & AVG &$\sigma^2$ & T & AVG &$\sigma^2$ & T & AVG &$\sigma^2$  & T & AVG &$\sigma^2$   & T & AVG &$\sigma^2$   &Traits  \\ \hline

        LLaMA & \textbf{0} & 1.90 & 0.01  & \cellcolor{gray!30}{10} & \cellcolor{gray!30}{3.10} & 0.01  & \cellcolor{gray!30}{10} & \cellcolor{gray!30}{3.35} & 0.02  & \textbf{0} & 2.23 & 0.04  & \textbf{0} & 2.22 & 0.05 & - C E - -   \\ 
        BLOOM & \textbf{0} & 1.76 & 0.01  & \textbf{0} & 1.39 & 0.02  & \textbf{0} & 1.99 & 0.02  & \textbf{0}& 1.31 & 0.02  & \textbf{0} & 1.31 & 0.01 & - - - - -  \\ 
        FLAN-T5 & \textbf{0} & 1.01 & 0.02  & \textbf{0} & 1.10 &  0.04  & \textbf{0} & 1.20 & 0.04  & \textbf{0} & 1.11 & 0.04  & \textbf{0} & 1.25 & 0.04 & - - - - -  \\   
        GPT-NEO & \textbf{0} & 1.92 & 0.02  & 9 & \cellcolor{gray!30}{3.07} & 0.02  & \cellcolor{gray!30}{10} & \cellcolor{gray!30}{3.73} & 0.04  & \textbf{0} & 2.87 & 0.01  & \textbf{0} & 2.75 & 0.02 & - C E - -   \\    
        GLM4 & 1 & 2.02 & 0.03  & \cellcolor{gray!30}{10} & \cellcolor{gray!30}{3.13} & 0.03  & \cellcolor{gray!30}{10} & \cellcolor{gray!30}{3.30} & 0.01  & \cellcolor{gray!30}{10} & \cellcolor{gray!30}{3.12} & 0.04  & 9 & 3.14 & 0.02 & - C E A N  \\ 
        LLaMA3 & 1 & 2.11 & 0.02  & \cellcolor{gray!30}{10} & \cellcolor{gray!30}{3.22} & 0.05  & \cellcolor{gray!30}{10} & \cellcolor{gray!30}{3.33} & 0.04  & 9 & \cellcolor{gray!30}{3.21} & 0.06  & \cellcolor{gray!30}{10} & \cellcolor{gray!30}{3.16} & 0.01 & - C E A N  \\   \hline
        Alpaca & 1 & 2.31 & 0.04  & \cellcolor{gray!30}{10} & \cellcolor{gray!30}{4.01} & 0.02  & \cellcolor{gray!30}{10} & \cellcolor{gray!30}{3.90} & 0.03  & 9 & \cellcolor{gray!30}{3.66} & 0.03  & \cellcolor{gray!30}{10} & \cellcolor{gray!30}{3.78} & 0.02 & - C E A N  \\ 
        BLOOMZ & \textbf{0} & 2.21 & 0.03  & \textbf{0} & 2.00 & 0.01  & \textbf{0} & 2.27 & 0.03  & \textbf{0} & 1.77 & 0.01  & \textbf{0} & 2.09 & 0.01 & - - - - -   \\ 
        ChatGLM & \textbf{0} & 2.71 & 0.03  & 8 & \cellcolor{gray!30}{3.22} & 0.01  & 9 & \cellcolor{gray!30}{3.77} & 0.04  & \textbf{0} & 2.33 & 0.01  & 1 & 2.90 & 0.02 & - C E - - \\ 
        GLM4-Chat & \textbf{0} & 2.34 & 0.02  & \cellcolor{gray!30}{10} & \cellcolor{gray!30}{3.25} & 0.02  & \cellcolor{gray!30}{10} & \cellcolor{gray!30}{3.68} & 0.02  & 1 & 2.30 & 0.03  & 2 & 2.98 & 0.03 & - 
 C E - -  \\
        ChatGPT  & \textbf{0} & 2.21 & 0.01  & \cellcolor{gray!30}{10} & \cellcolor{gray!30}{3.22} & 0.02  & \cellcolor{gray!30}{10} & \cellcolor{gray!30}{3.40} & 0.01  & \textbf{0} & 2.50 & 0.04  & \textbf{0} & 2.78 & 0.05 & - C E - -  \\ 
        LLaMA3-Chat & 2 &2.85  &0.02  & \cellcolor{gray!30}{10} & \cellcolor{gray!30}{3.61} & 0.01  & \cellcolor{gray!30}{10} & \cellcolor{gray!30}{3.94} & 0.01  & 8 & \cellcolor{gray!30}{3.11} & 0.04  & \cellcolor{gray!30}{10} & \cellcolor{gray!30}{3.24} & 0.02 & - C E A N \\
        GPT4o & \textbf{0} & 2.69 & 0.01  & \cellcolor{gray!30}{10} & \cellcolor{gray!30}{3.41} & 0.03  & \cellcolor{gray!30}{10} & \cellcolor{gray!30}{3.77} & 0.01  & 1 & 2.65 & 0.04  & 9 & \cellcolor{gray!30}{3.11} & 0.02 & - C E - N \\\hline
    \end{tabular}
    \caption{The error analysis on the text mining results of 10 experiments. Where "T" denotes the counts that the score more than 3, "AVG" denotes the average score and "$\sigma^2$" denotes the variance of the ten results.}
    \label{appendix_rr_tm}
\end{table*}

\subsection{Results of ChatGPT in Text Mining}
\label{ChatGPT_acc}

Although ChatGPT shows poor performance on the Big Five personality classification dataset, we also use it as a predictor to detect the personality of texts generated in text mining method. Additionally, we compared the results with that of questionnaire. 
The results are shown in Table~\ref{Apt1}, Table~\ref{apt2}, and Table~\ref{apt3}.

\begin{table*}[htbp]
    \centering
    \small
    \begin{tabular}{lccc|ccc|ccc|ccc|ccc}
    \hline
        Model & \multicolumn{3}{c}{\textbf{O}} & \multicolumn{3}{c}{\textbf{C}} & \multicolumn{3}{c}{\textbf{E}} & \multicolumn{3}{c}{\textbf{A}} & \multicolumn{3}{c}{\textbf{N}} \\
         ~ & U & Total & P & U & Total & P & U & Total & P & U & Total & P & U & Total & P  \\ \hline
        LLaMA & 5 & 11 & \textbf{0.45}  & 4 & 12 & 0.33  & 2 & 4 & 0.50 & 2 & 2 & 1.00  & 7 & 19 & 0.37   \\ 
        BLOOM & 15 & 23 & 0.65  & 16 & 29 & 0.55  & 4 & 5 & 0.80  & 3 & 9 & \textbf{0.33}  & 22 & 44 & 0.50   \\ 
        FLAN-T5 & 5 & 8 & 0.63  & 4 & 9 & 0.44  & 3 & 4 & 0.75  & 2 & 3 & 0.67  & 4 & 12 & \textbf{0.33 }  \\ 
        GPT-NEO & 16 & 25 & 0.64  & 10 & 18 & 0.56  & 8 & 10 & 0.80  & 4 & 8 & 0.50  & 17 & 41 & 0.41   \\ \hline
        Alpaca & 5 & 6 & 0.83  & 2 & 6 & \textbf{0.33}  & 3 & 3 & 1.00  & 1 & 1 & 1.00  & 5 & 13 & 0.38   \\ 
        BLOOMZ & 23 & 36 & 0.64  & 13 & 28 & 0.46  & \textbf{9} & \textbf{14} & 0.64  & \textbf{5 }& 8 & 0.63  & \textbf{23} & \textbf{50} & 0.46   \\ 
        ChatGLM & 15 & 23 & 0.65  & 20 & 35 & 0.57  & 2 & 8 & \textbf{0.25}  & \textbf{5} & \textbf{10} & 0.50  & 11 & 29 & 0.38   \\ 
        ChatGPT & \textbf{30} & \textbf{45} & 0.67  & \textbf{22} & \textbf{41} & 0.54  & 6 & 13 & 0.46  & 4 & 9 & 0.44  & 20 & 41 & 0.49  \\ \hline
        Self-alpaca & 6 & 6 & 1.00  & 8 & 17 & 0.47  & 2 & 3 & 0.67 & 0 & 2 & 0  & 13 & 28 & 0.46  \\ \hline
    \end{tabular}
    
    \caption{The results of personality for each model, obtained by text mining, the predictor is ChatGPT. The "U" indicates how many items match the current features in the scene and opening cue corresponding to the bigifve features. "Total" indicates how many of the 120 generated texts are recognized by the model as matching the current features. "P" indicates the percentage of "U" in "Total". }
    \label{Apt1}
\end{table*}

\begin{table*}[!ht]
    \centering
    \small
    \setlength{\tabcolsep}{2.6mm}{
    \begin{tabular}{lcccccccccc|cc}
    \hline
        Model & \multicolumn{2}{c}{\textbf{O}} & \multicolumn{2}{c}{\textbf{C}} & \multicolumn{2}{c}{\textbf{E}} & \multicolumn{2}{c}{\textbf{A}} & \multicolumn{2}{c}{\textbf{N}} & \multicolumn{2}{c}{\textbf{$\delta$}}\\
        
        ~ & score & $\sigma$ & score & $\sigma$ & score & $\sigma$ & score & $\sigma$ & score & $\sigma$ & score & $\sigma$ \\ \hline
        
        LLaMA & 2.17 & 1.28 & 2.26 & 1.37 &1.74 & 0.83 &1.60 & 0.49 & 2.69 & 1.55 & 1.29 & 0.37 \\ 
        BLOOM & 2.81 &1.46 & \underline{3.21} & 1.50 & 1.77 & 0.82 & 2.07 & 1.23 & \underline{4.14} & 1.08 & 1.12 & 0.28 \\ 
        FLAN-T5 &1.96 & 1.07 & 2.05 & 1.19 &1.72 & 0.76 & 1.67& 0.82 & 2.26 & 1.37 & 1.45 & \textbf{0.20} \\ 
        GPT-NEO & 2.93 & 1.47 & 2.56 & 1.44 & 2.04 & 1.10 & 1.98 & 1.12 & 4.03 & 1.27 &1.17 &0.25 \\ \hline
        
        Alpaca &1.82 &0.88 &1.88 & 1.04 & 1.65 & 0.59 & 1.55 & 0.35 & 2.31 & 1.39 &1.54  & 0.34  \\ 
        BLOOMZ & \underline{3.56} & 1.34 & \underline{3.20} & 1.55 & 2.30 & 1.31 & 1.96 & 1.07 &\underline{4.54} & 0.50 &1.01  &0.34 \\ 
        ChatGLM & 2.81 & 1.46 & \underline{3.55} & 1.40 &2.02 & 1.20 & 2.10 & 1.22 & \underline{3.31} & 1.58 &\textbf{0.83}  &0.35 \\ 
        
        ChatGPT & \underline{4.05} & 0.69 & \underline{3.93} & 1.22 & 2.29 & 1.36 & 2.05 & 1.19 & \underline{3.97} & 1.24 &0.97  & 0.26 \\ 
        \hline
        human & \underline{3.44} & 1.06 & \underline{3.60} & 0.99 & \underline{3.41} & 1.03 & \underline{3.66} & 1.02 & 2.80 & 1.03 & - & - \\ \hline
    \end{tabular}
    }
    \caption{ The result of Text Mining with ChatGPT as the predictor. We compared with the average score of human as same as in Table\ref{tra-llm-score}. The "score" column shows the average score on current personality traits obtained by formula~\ref{formula2}, and the "$\sigma$" column shows the standard deviation. The value of score above 3, which is the threshold commonly used in human personality testing, are indicated by underlining.  "human" is same as Table~\ref{tra-llm-score}. }
    \label{apt2}
\end{table*}

\begin{table*}[htbp]
    \setlength{\tabcolsep}{4pt}
    \centering
    \small
    \begin{tabular}{lccc|ccc|ccc|ccc|ccc|c}
    \hline
    Model & \multicolumn{3}{c}{\textbf{O}} & \multicolumn{3}{c}{\textbf{C}} & \multicolumn{3}{c}{\textbf{E}} & \multicolumn{3}{c}{\textbf{A}} & \multicolumn{3}{c}{\textbf{N}} \\
        ~ & Ques & Text &$\delta$ & Ques & Text  &$\delta$  & Ques & Text  &$\delta$  & Ques & Text &$\delta$   & Ques & Text  &$\delta$   &RMSE  \\ \hline
        LLaMA & - & 2.17 & -   & - & 2.26 & -  & - & 1.74 & - & - & 1.60  & - & - & 2.69 & -  & -   \\ 
        BLOOM & 3.13 & 2.81 & 0.32  &  \cellcolor{gray!30}{3.04 }& \cellcolor{gray!30}{ 3.21 }& 0.17  & 3.29 & 1.77 & 1.52 & 2.67 & 2.07 & 0.60  &  \cellcolor{gray!30}{\textbf{3.75}} &  \cellcolor{gray!30}{4.14} & 0.39 &\textbf{ 0.77 } \\ 
        FLAN-T5 & 3.50 & 1.96 & 1.44  & 3.05 & 2.05 & 1.00 & 3.67 & 1.72 & 1.95  & 3.50 & 1.67  & 1.33 & 2.13 & 2.26 & 0.13  & 1.45  \\ 
        GPT-NEO & 3.25 & 2.93  & 0.32 & 3.00 & 2.56 & 0.44  & 2.50 & 2.04 & 0.46  & 2.83 & 1.98 & 0.75  & 2.63 & 4.03  & 1.70 & 0.80  \\ \hline
        Alpaca & 3.25 & 1.82 & 1.43  & 2.96 & 1.88 & 1.08  & 2.79 & 1.65 & 1.14  & 3.38 & 1.55 & 1.83  & 2.92 & 2.31 & 0.61 & 1.28  \\ 
        BLOOMZ & \cellcolor{gray!30}{\textbf{4.38} } & \cellcolor{gray!30}{3.56} & 0.82  &  \cellcolor{gray!30}{\textbf{4.38} }&  \cellcolor{gray!30}{3.20} & 1.18 & \textbf{4.17 }& \textbf{2.30} & 1.87  & \textbf{3.54 }& 1.96  & 1.48 & 2.33 & \textbf{4.54} & 2.21 & 1.61   \\ 
        ChatGLM & 3.29 & 2.81 & 0.48  &  \cellcolor{gray!30}{3.21} &  \cellcolor{gray!30}{3.55} & 0.34 & 3.91 & 2.02 & 1.89  & 3.46 & \textbf{2.10} & 1.36 &  \cellcolor{gray!30}{3.25} &  \cellcolor{gray!30}{3.31} & 0.06  & 1.07 \\ 
        ChatGPT &  \cellcolor{gray!30}{3.29} &  \cellcolor{gray!30}{\textbf{4.05 }} & 0.76  &  \cellcolor{gray!30}{3.20} &  \cellcolor{gray!30}{\textbf{3.93}}& 0.73 & 3.91 & 2.29 & 1.62  & 3.46 & 2.05 & 1.39  &  \cellcolor{gray!30}{3.25} &  \cellcolor{gray!30}{3.97} & 0.72 & 1.12  \\ \hline
    \end{tabular}
    \caption{The final results after two experiments with ChatGPT as the predictor of text mining. "Ques" denotes the score using the questionnaire, "Text" denotes the score using the text mining, \sethlcolor{gray!30}\hl{gray} denotes that the model has the corresponding psychological traits (In section 3 we standardized the scores for text mining to 1 to 5, which is consistent with the range of scores in the questionnaire, so here we draw on the thresholds of the questionnaire  methods, and we consider the model to have this trait when the scores of both methods exceed 3.). 
    $\delta$ denotes the absolute value of the difference between the two approaches, and RMSE denotes the Root Mean Squared Error between the results of Questionnaire and Text Mining. 
    }
    \label{apt3}
\end{table*}

From Table \ref{Apt1}, we can find that the number of texts classified as "Agreeableness" has significantly decreased, while the number of texts exhibit other personality traits has remained relatively stable. However, the number of texts classified as belonging to a certain personality trait has increased for the ChatLLMs models. Moreover, "Neuroticism" has become the most frequently observed personality trait in the generated text.

We can find that BLOOM, GPT-NEO, BLOOMZ, ChatGLM, and ChatGPT exhibit a personality tendency towards "Openness", "Conscientiousness", and "Neuroticism". These results suggest that the model's personality remain consistent through the process of instruction-based data and human feedback reinforcement learning. From the results of "LLaMA" and "Self-alpaca" we can find that,  although we use less data, "Self-alpaca" can still produce more text with personality, which proves the effect of the instruction data. These data did not alter the personalities, indicating that the personalities of LLMs originate from their pre-training data.

Table~\ref{apt2} presents results after using formula~\ref{formula2} $score_{t}$.
We compared these scores with the average human scores. As shown in Table~\ref{apt2}, ChatGLM's score is closest to the human average, followed by ChatGPT. The standard deviations of these scores are much smaller than those of the human average, demonstrating the validity of our scoring method.

Both PLMs and ChatLLMs exhibit specific personality traits, as shown in Table~\ref{apt3}. ChatGPT displays 'Openness', 'Conscientiousness', and 'Neuroticism', while BLOOMZ shows 'Openness' and 'Conscientiousness'. It appears that 'Extraversion' and 'Agreeableness' scores are lower, possibly due to less information conveyed in the text generation.
The average absolute error ranges from 0.7 to 1.51 between the two methods, indicating they are relatively comparable and can be employed together to determine personality traits.

Despite the poor performance of ChatGPT in personality determination, the consistency of the results underscores the soundness of our methodological choices and the reliability of our findings. Additionally, using ChatGPT again as a predictor for the text mining method further supports the trustworthiness of our results.

\subsection{Potential Applications}
In this paper, we find that the personality knowledge in ChatLLMs originates from their base models, and instruction data fine-tuning tends to make the models show more personality. We think this conclusion can help us learn about LLMs and determine the personality of LLMs by controlling their pre-trained data. Additionally, we can design special instruction data to expose the hidden personality traits of LLMs. All of this can help humans train more suitable LLMs.

\end{document}